\documentclass[letterpaper, 10 pt, conference]{ieeeconf}  % Comment this line out if you need a4paper

\IEEEoverridecommandlockouts                              % This command is only needed if 
\usepackage{graphicx} % for pdf, bitmapped graphics files
\usepackage{amsmath} % assumes amsmath package installed
\usepackage{amssymb}  % assumes amsmath package installed
\usepackage{booktabs} % for professional-quality tables
\usepackage{multirow} % for tables with merged cells
\usepackage{url} % for URLs
\usepackage{subfig}
\usepackage{textcomp}
\usepackage{subcaption} 
\usepackage[utf8]{inputenc}
\title{\LARGE \bf
Technical Report for ICRA 2025 GOOSE 3D Semantic Segmentation Challenge: Adaptive Point Cloud Understanding for Heterogeneous Robotic Systems
}

\author{Xiaoya Zhang$^{1}$% <-this % stops a space
\thanks{$^{1}$EARTHBRAIN Ltd., Japan.
        Email: {\ xiaoya\_zhang@earthbrain.com}}%
}
\usepackage{fancyhdr}

\fancypagestyle{withfooter}{
  
  \fancyfoot[C]{\footnotesize Winners of the GOOSE 3D Semantic Segmentation Challenge at the IEEE ICRA Workshop on Field Robotics 2025}
}
\begin{document}

\maketitle
\thispagestyle{withfooter}
\pagestyle{withfooter}

%%%%%%%%%%%%%%%%%%%%%%%%%%%%%%%%%%%%%%%%%%%%%%%%%%%%%%%%%%%%%%%%%%%%%%%%%%%%%%%%
\begin{abstract}
This technical report presents the implementation details of the winning solution for the ICRA 2025 GOOSE 3D Semantic Segmentation Challenge. This challenge focuses on semantic segmentation of 3D point clouds from diverse unstructured outdoor environments collected from multiple robotic platforms. This problem was addressed by implementing Point Prompt Tuning (PPT) integrated with Point Transformer v3 (PTv3) backbone, enabling adaptive processing of heterogeneous LiDAR data through platform-specific conditioning and cross-dataset class alignment strategies. The model is trained without requiring additional external data. As a result, this approach achieved substantial performance improvements with mIoU increases of up to 22.59\% on challenging platforms compared to the baseline PTv3 model, demonstrating the effectiveness of adaptive point cloud understanding for field robotics applications.

\end{abstract}

%%%%%%%%%%%%%%%%%%%%%%%%%%%%%%%%%%%%%%%%%%%%%%%%%%%%%%%%%%%%%%%%%%%%%%%%%%%%%%%%
\section{INTRODUCTION}

The growing availability of 3D point cloud data from LiDAR sensors requires effective representation learning to transform raw spatial data into semantic information for autonomous systems and robotics navigation. Large-scale learning is crucial for developing robust models that generalize well across diverse real-world scenarios and capture complex environmental details. High-capacity models are able to learn nuanced patterns, while pre-training on existing datasets can reduce reliance on manual annotation, ultimately improving field robotics' ability to perceive and interact with complex, unstructured 3D environments.

The German Outdoor and Offroad Dataset(GOOSE) 3D Semantic Segmentation Challenge, hosted in conjunction with the Workshop on Field Robotics at ICRA 2025, aims to drive innovation in 3D perception for field robotics applications. The challenge is based on GOOSE\cite{goose} and GOOSE-Ex\cite{gooseex} LiDAR datasets which capture diverse unstructured outdoor scenes collected from multiple robotic platforms. Participants are challenged to develop semantic segmentation models for LiDAR point scans, with submissions evaluated on withheld test data using mean Intersection over Union (mIoU). A baseline using Point Transformer V3(PTv3)\cite{ptv3} is provided to establish performance benchmarks.

3D semantic segmentation is particularly important for robots operating in unstructured outdoor environments because it enables granular understanding of surroundings, which is fundamental for safe navigation, effective path planning, environmental interaction, and autonomous decision-making in complex and unpredictable settings where pre-defined maps may be unavailable or unreliable. The diverse robotic platforms and sensor configurations of GOOSE and GOOSE-Ex datasets, present significant challenges for traditional segmentation approaches that often struggle with domain shift and negative transfer effects when training across such varied data sources.

For effective segmentation on these combined datasets, models must excel at handling domain shifts. Given the diverse robotic platforms inherent in these datasets, adaptive architectures are essential. Experiments have demonstrated that Point Prompt Tuning (PPT)\cite{ppt} framework which combines a strong general-purpose point cloud backbone with mechanisms for domain-specific adaptation, is optimally suited to handle the heterogeneity of the combined GOOSE and GOOSE-Ex datasets. This approach enables learning of common semantic features while maintaining sensitivity to unique platform-specific characteristics, achieving substantial performance improvements with mIoU increases of up to 22.59\% on challenging platforms compared to the baseline PTv3 model.
\begin{table*}[htpb!]
	\vspace{1em}
	\centering
	\caption{
		Dataset Statistics and Platform Characteristics
	}
	\renewcommand{\arraystretch}{1.1} % slightly increases the row height to give some space after the dashed line
	\begin{tabular}{@{}lccccr@{}}
		\toprule
		Platform & LiDAR Setup & Training & Validation & Test & Avg Points per scan \\ \midrule
		MuCAR-3 & 1×128 ring & 7\,719 & 961 & 1\,211 & 186\,290 \\
		ALICE & 3×64 ring + 1×128 ring & 2\,164 & 192 & 444 & 278\,628 \\
		Spot & 1×64 ring & 1\,825 & 215 & 160 & 89\,470 \\
		\bottomrule
	\end{tabular}
	\label{tab:dataset_stats}
\end{table*}

\begin{figure*}[ht!]
\centering
\includegraphics[width=6in]{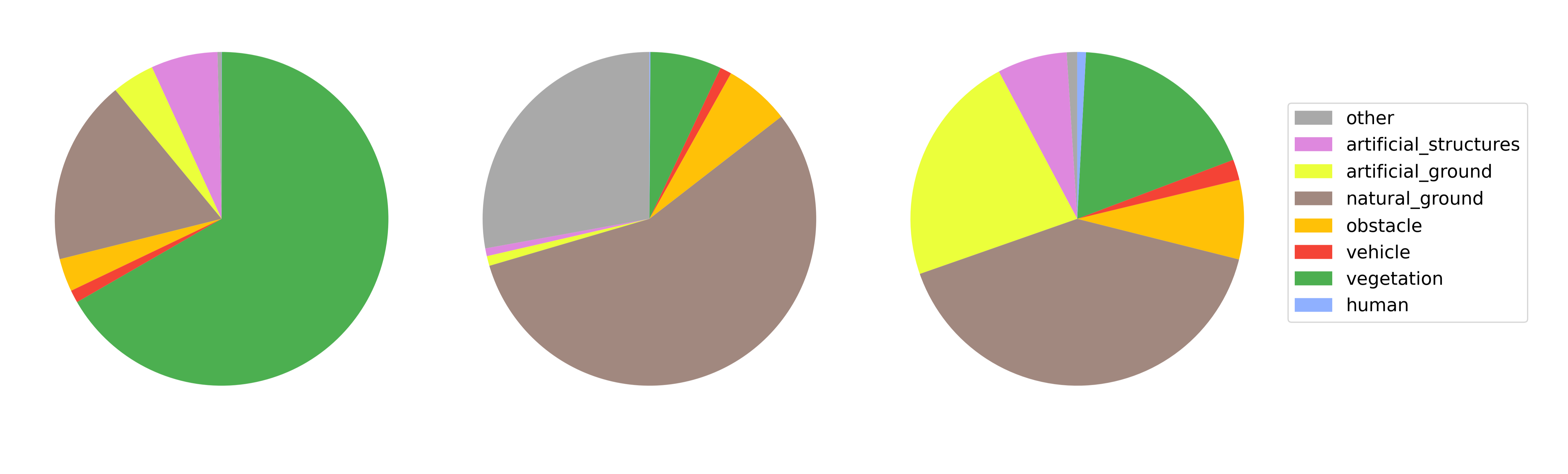}
\caption{Label distribution across LiDAR scans from MuCAR-3(left), ALICE(middile), and Spot(right) platforms.}
\label{pie_chart}
\end{figure*}

\section{RELATED WORK}

Recent advancement in 3D semantic segmentation have produced strong baseline models with notable strengths. Architectures like MinkUNet\cite{minkunet} and Point Transformers series\cite{pt}\cite{ptv2}\cite{ptv3} effectively capture geometric features from 3D point clouds using sparse convolutions or attention-based mechanisms. These models provide a solid foundation for semantic understanding in structured environment such as urban road scenes.

Additionally, fusion techniques that combine 2D image features with 3D point cloud data have introduced significant improvements. By incorporating RGB or RGB-D data, models like 2DPASS\cite{2dpass} and LCPS\cite{lcps} enhance semantic richness, especially in sparse LiDAR scenes. More recently, foundation model distillation methods such as DITR\cite{DITR} and D-DITR\cite{DITR} have leveraged powerful 2D vision foundation models (VFMs) such as DINOv2\cite{dinov2} to transfer semantic knowledge into 3D backbones, yielding improved performance with limited or no labeled 3D data. Furthermore, self-supervised\cite{selfsuper} and contrastive learning strategies\cite{contrast} have helped address the scarcity of annotations by aligning representations across modalities.

Despite these strengths, several critical limitations remain—particularly when extending models to generalize across multiple types of unstructured outdoor terrain. One major issue is that many existing models are trained and tested on structured urban datasets, such as SemanticKITTI\cite{kitti} or nuScenes\cite{nusense}, which introduces a strong dataset-specific bias, but when applied to unstructured natural terrains the performance degrades due to a lack of diverse representation during training.

Another significant challenge is negative transfer during multi-dataset training. Simply merging diverse datasets for joint training can result in reduced performance on individual datasets, and this happens because of differences in point density, scene complexity, and semantic definitions across datasets. Furthermore, most datasets have different label spaces, making it difficult to unify semantic categories. This mismatch leads to inefficient training and limited transferability.

\textbf{Point Prompt Training (PPT)} was proposed by Wu et al.\cite{ppt} to directly address these issues. It introduces two key innovations: \textit{Prompt-driven Normalization} and \textit{Categorical Alignment}. The former allows a single model to adapt to the specific context of different datasets using learnable domain-specific prompts. This mitigates the problem of negative transfer by helping the model distinguish between dataset distributions while maintaining shared weights. The latter component aligns category semantics across datasets by linear projection or by embedding class labels using language model such as CLIP\cite{clip}, which allows the model to reconcile similar categories even if they differ in naming or granularity.

Together, these mechanisms enable PPT to train a single model across multiple datasets of different domain gaps,while still maintaining or even improving performance. Moreover, PPT models do not require images at inference, making them more applicable in real-world unstructured terrain where usually only 3D LiDAR scans are available.

\section{DATASET DESCRIPTION}

The combined GOOSE\cite{goose} and GOOSE-Ex\cite{gooseex} datasets consist of 13,076 labeled point cloud across three robotic platforms. The dataset exhibits significant heterogeneity in both scale and semantic distribution across platforms. Table~\ref{tab:dataset_stats} summarizes the dataset characteristics.

The challenge uses 7 superclasses (artificial structures, artificial ground, natural ground, obstacle, vehicle, vegetation and human) derived from the original 64 classes.

The GOOSE\cite{goose} dataset aims to address the critical need for large-scale annotated 3D LiDAR data from challenging unstructured outdoor environments and diverse weather conditions typical of field robotics operations.The data was recorded on the UniBw Munich research vehicle MuCAR-3\cite{mucar3}. 
GOOSE-Ex\cite{gooseex} dataset significantly extends this foundation by introducing four specialized environmental contexts: generic off-road and industrial regions, landfill environments representing typical excavator operating conditions, quarry settings with complex surface geometries for large machinery operations, and construction sites featuring heavy equipment working areas. The main recording platforms include a Liebherr R924 track excavator ALICE\cite{alice}, and a Boston Dynamics Spot robot with two different sensor setups.

The datasets show pronounced platform-specific semantic biases as illustrated in figure\ref{pie_chart}. MuCAR-3 is dominated by vegetation (66.65\%) and natural ground (17.91\%), reflecting forest and natural terrain environments. ALICE exhibits a more balanced distribution with natural ground (55.95\%) and a significant proportion of unlabeled/other points (27.85\%), suggesting more complex mixed environments. Spot shows the most diverse semantic distribution, with natural ground (40.73\%), artificial ground (22.57\%), and vegetation (18.40\%) being prominent, indicating operation in more structured outdoor environments.

Severe class imbalances are evident across all platforms, particularly for human instances (0.03-0.84\%) and vehicles (1.12-1.99\%). Obstacle class shows platform-specific variations(3.20-7.68\%) , while artificial structures representation varies significantly (0.75-6.77\%), reflecting the diverse operational contexts of each robotic platform.

% MuCAR-3 Summary Table
\begin{table*}[htpb!]
	\vspace{1em}
	\centering
	\caption{Per-Class IoU/Accuracy and Overall Validation Metrics for MuCAR-3}
	\resizebox{\textwidth}{!}{%
		\renewcommand{\arraystretch}{1.1}
		\begin{tabular}{@{}lcccccccccc@{}}
			\toprule
			Models/Classes & Artificial Structures & Artificial Ground & Natural Ground & Obstacle & Vehicle & Vegetation & Human & mIoU & mAcc & allAcc \\ \midrule
			PTv3(baseline) & 0.8639 / 0.8923 & 0.7438 / 0.8114 & 0.8025 / 0.9220 & 0.6450 / 0.7939 & 0.9082 / 0.9311 & 0.9191 / 0.9510 & 0.7558 / 0.8253 & 0.8295 & 0.8909 & 0.9295 \\
			PPT-LA & 0.8656 / 0.9180 & 0.8133 / 0.9420 & 0.8208 / 0.8842 & 0.6378 / 0.7202 & 0.9109 / 0.9445 & 0.9272 / 0.9715 & 0.7536 / 0.8522 & 0.8411 & 0.9041 & 0.9375 \\
			PPT-DA & \textbf{0.8964} / 0.9337 & \textbf{0.8716} / 0.9291 & \textbf{0.8512} / 0.9283 & \textbf{0.6952} / 0.7840 & \textbf{0.9218} / 0.9358 & \textbf{0.9382} / 0.9689 & \textbf{0.7908} / 0.8686 & \textbf{0.8706} & \textbf{0.9185} & \textbf{0.9488} \\
			\bottomrule
		\end{tabular}
	}
	\label{tab:mucar3_summary}
\end{table*}

% Alice Summary Table
\begin{table*}[htpb!]
	\vspace{1em}
	\centering
	\caption{Per-Class IoU/Accuracy and Overall Validation Metrics for ALICE}
	\resizebox{\textwidth}{!}{%
		\renewcommand{\arraystretch}{1.1}
		\begin{tabular}{@{}lcccccccccc@{}}
			\toprule
			Models/Classes & Artificial Structures & Artificial Ground & Natural Ground & Obstacle & Vehicle & Vegetation & Human & mIoU & mAcc & allAcc \\ \midrule
			PTv3(baseline) & 0.9069 / 0.9679 & 0.0333 / 0.0362 & \textbf{0.8852} / 0.9931 & 0.0671 / 0.0697 & 0.7793 / 0.8006 & 0.8569 / 0.8950 & 0.7850 / 0.7951 & 0.6631 & 0.6947 & 0.9178 \\
			PPT-LA & \textbf{0.9136} / 0.9578 & \textbf{0.4195} / 0.4470 & 0.8831 / 0.9762 & 0.1008 / 0.1229 & 0.8881 / 0.9711 & 0.9022 / 0.9283 & \textbf{0.8721} / 0.8879 & \textbf{0.7474} & \textbf{0.7864} & \textbf{0.9197} \\
			PPT-DA & 0.9011 / 0.9741 & 0.2069 / 0.2121 & 0.8753 / 0.9716 & \textbf{0.1142} / 0.1459 & \textbf{0.9116} / 0.9720 & \textbf{0.9061} / 0.9209 & 0.8606 / 0.8734 & 0.7220 & 0.7587 & 0.9145 \\
			\bottomrule
		\end{tabular}
	}
	\label{tab:alice_summary}
\end{table*}

% Spot Summary Table
\begin{table*}[htpb!]
	\vspace{1em}
	\centering
	\caption{Per-Class IoU/Accuracy and Overall Validation Metrics for Spot}
	\resizebox{\textwidth}{!}{%
		\renewcommand{\arraystretch}{1.1}
		\begin{tabular}{@{}lcccccccccc@{}}
			\toprule
			Models/Classes & Artificial Structures & Artificial Ground & Natural Ground & Obstacle & Vehicle & Vegetation & Human & mIoU & mAcc & allAcc \\ \midrule
			PTv3(baseline) & 0.5539 / 0.8570 & 0.7263 / 0.8628 & 0.6514 / 0.8765 & 0.3175 / 0.3452 & 0.7973 / 0.8330 & 0.8542 / 0.8904 & 0.8056 / 0.8094 & 0.7087 & 0.8093 & 0.7969 \\
			PPT-LA & \textbf{0.8627} / 0.8946 & \textbf{0.8395} / 0.9251 & \textbf{0.7956} / 0.8780 & 0.7277 / 0.8354 & 0.8948 / 0.9501 & 0.8581 / 0.9328 & \textbf{0.9716} / 0.9784 & \textbf{0.8688} & 0.9243 & 0.8959 \\
			PPT-DA & 0.8235 / 0.9212 & 0.8371 / 0.9505 & 0.7909 / 0.8498 & \textbf{0.7366} / 0.8662 & \textbf{0.8895} / 0.9278 & \textbf{0.8740} / 0.9422 & 0.9710 / 0.9763 & 0.8653 & \textbf{0.9292} & \textbf{0.8971} \\
			\bottomrule
		\end{tabular}
	}
	\label{tab:spot_summary}
\end{table*}

\section{METHODOLOGY}

\subsection{Network Architecture Overview}

 An open-source implementation of a unified framework that combines Point Transformer v3 (PTv3) with Point Prompt Tuning (PPT) is adopted to address the fundamental challenge of learning from diverse 3D point cloud data while maintaining consistent semantic understanding across different robotic platforms. This approach enables effective cross-platform knowledge transfer while preserving platform-specific adaptation capabilities.

\subsection{PTv3 Backbone}

This network employs PTv3 as the primary feature extractor, leveraging its state-of-the-art transformer-based architecture specifically designed for point cloud processing. PTv3 excels at capturing complex local and global geometric relationships through attention mechanisms and hierarchical feature extraction, providing a robust foundation for multi-platform point cloud understanding.

\subsection{Point Prompt Tuning Framework}

The PPT framework addresses heterogeneous data processing through two core mechanisms: data-driven context adaptation and cross-dataset class alignment.

\subsubsection{Data-Driven Context Adaptation}

To handle platform-specific characteristics, conditional normalization was implemented that adapts model behavior based on data source origin: The backbone incorporates decoupled normalization layers with platform-specific conditions (``car'', ``alice'', ``spot''). This enables parameter adjustment based on input source characteristics, allowing the shared backbone to process features optimally for each platform's unique sensor and environmental properties.

\subsubsection{Cross-Dataset Class Alignment Strategies}

To unify diverse class taxonomies while preserving semantic consistency, two complementary alignment approaches are implemented:

\textbf{Language-Driven Alignment (PPT-LA):} This approach leverages pre-trained CLIP (ViT-B/16) to create semantic bridges between different label spaces. Class names are converted to text prompts to generate text embeddings that serve as semantic anchors. Point features are projected into the CLIP embedding space, enabling segmentation through feature-to-text similarity comparison.

\textbf{Decoupled Alignment (PPT-DA):} This strategy employs separate segmentation heads for each data condition while maintaining shared backbone features. Each decoupled pathway is trained to predict the same target superclass set, eliminating the need for explicit cross-dataset class mapping at the output layer while leveraging condition-aware backbone representations.

\subsection{Training Strategy}

The model is trained with data from all platforms. Each sample is processed through the condition-aware backbone, with the alignment strategy determining the final segmentation pathway. This unified training enables the model to learn from collective knowledge across diverse datasets while preserving platform-specific adaptation capabilities.

\section{EXPERIMENTS AND RESULTS}

\subsection{Experimental Setup}

This PPT+PTv3 approach is trained and evaluated on the combined GOOSE and GOOSE-Ex datasets, including data captured by three robotic platforms: MuCAR-3, ALICE, and Spot (Spot v1 and Spot v2 merged). The datasets were evaluated separately to analyze platform-specific performance and the effectiveness of cross-platform adaptation.

\textbf{Evaluation Metrics:} Primary metrics include mean Intersection over Union (mIoU), mean accuracy (mAcc), and overall accuracy (allAcc) across 7 superclasses.

\textbf{Implementation Details:}
\begin{itemize}
\item \textbf{Input Channels:} 4 (coordinates + intensity)
\item \textbf{Batch Size:} 4
\item \textbf{Epochs:} 50
\item \textbf{Optimizer:} AdamW\cite{adamw} with learning rate 0.0008, weight decay 0.005.
\item \textbf{Scheduler:} OneCycleLR\cite{lr} with different rates for backbone (0.00005) and other components (0.0008).
\item \textbf{Loss Functions:} CrossEntropyLoss + LovaszLoss\cite{loss} (equally weighted )
\item \textbf{Backbone:} PTv3
\item \textbf{Prompt Tuning:} PPT with dataset-specific conditions for conditional normalization and feature adaptation.
\item \textbf{Alignment Strategies:}
  \begin{itemize}
  \item \textit{PPT-LA (Language-Driven):} Uses CLIP-based text embeddings for semantic class alignment.
  \item \textit{PPT-DA (Decoupled):} Uses separate segmentation heads for each platform condition.
  \end{itemize}
\item \textbf{Training:} Multi-dataset training with concatenated datasets using platform-specific conditioning.
\end{itemize}
\begin{figure*}[ht!]
\centering
\includegraphics[width=6in]{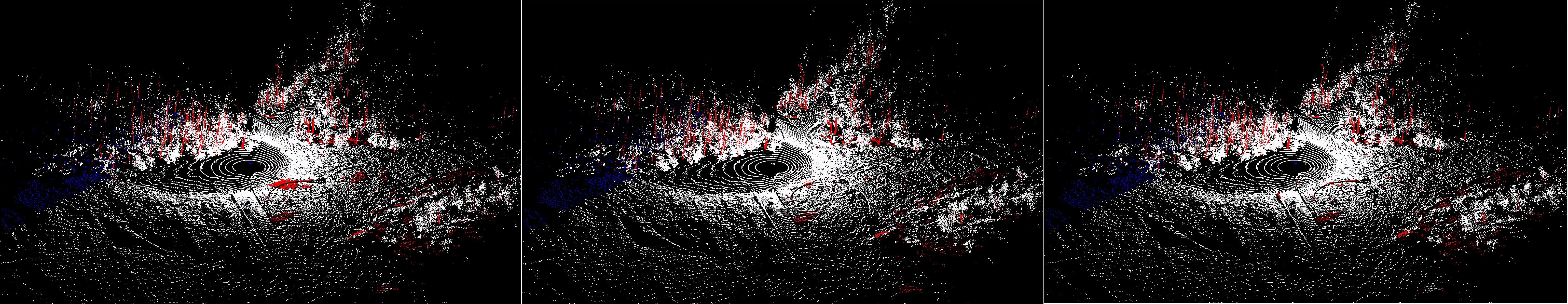}
\caption{Samples MuCAR-3 Point Cloud Segmentation Evaluation Results from PT-v3(left), PPT-LA(middle), PPT-DA(right). False Negatives (FN) are shown in red, False Positives (FP) in blue, and True Positives (correct predictions) in white }
\label{car_sample}
\end{figure*}
\begin{figure*}[ht!]
\centering
\includegraphics[width=6in]{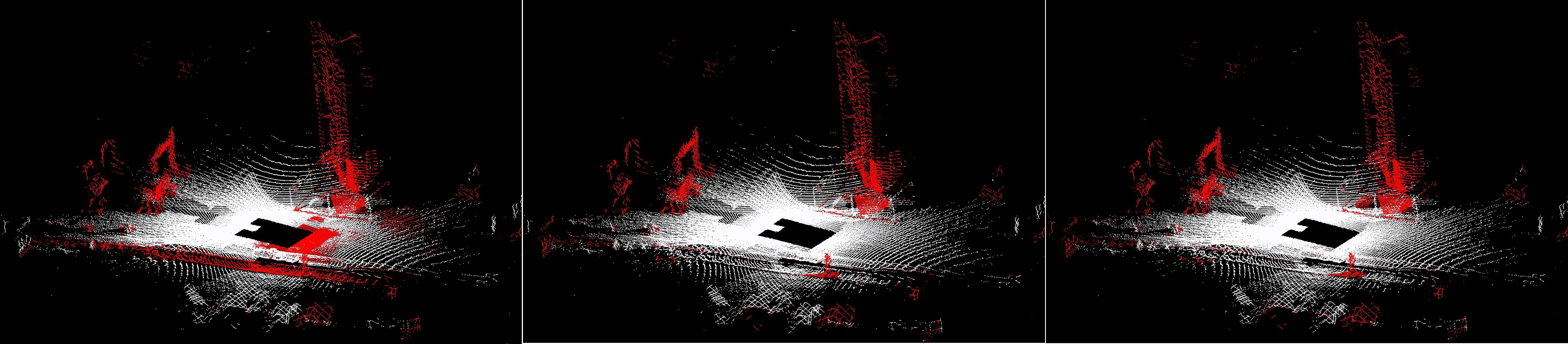}
\caption{Same as Figure\ref{car_sample} but for ALICE.}
\label{alice_sample}
\end{figure*}
\begin{figure*}[ht!]
\centering
\includegraphics[width=6in]{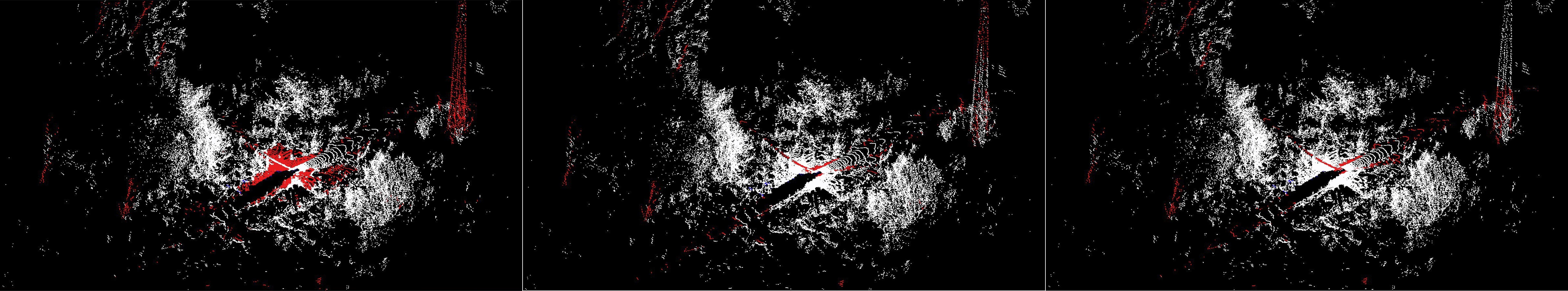}
\caption{Same as Figure\ref{car_sample} but for Spot}
\label{spot_sample}
\end{figure*}

\section{DISCUSSION}

\subsection{Summary of Findings}

The results demonstrate that PTv3+PPT is highly effective across different platforms, with substantial improvements on ALICE (+12.71\% mIoU) and Spot (+22.59\% mIoU), while showing more modest but consistent gains on MuCAR-3 (+4.95\% mIoU). The PPT-DA approach performs best on MuCAR-3, while PPT-LA excels on ALICE and Spot platforms.

\subsection{Analysis of Platform-Specific Performance}

\textbf{MuCAR-3 Platform:} The baseline PTv3 already achieves strong performance, indicating that this platform's data characteristics align well with standard architectures. However, PPT-DA still provides meaningful improvements, particularly in artificial ground and obstacle segmentation.

\textbf{ALICE Platform:} The baseline shows severe deficiencies in artificial ground and obstacle classes. PPT-LA dramatically improves artificial ground segmentation, demonstrating PPT's ability to leverage cross-platform knowledge through CLIP-based class understanding.

\textbf{Spot Platform:} Both PPT methods show substantial improvements across all classes. The consistent improvements across diverse semantic classes indicate that Spot data benefits significantly from multi-platform training and adaptive feature extraction.

\subsection{Effectiveness of Alignment Strategies}

\textbf{Language-Driven Alignment (PPT-LA)} demonstrates superior performance on challenging platforms (ALICE and Spot), particularly excelling in cross-domain knowledge transfer. The semantic grounding provided by CLIP embeddings facilitates better understanding of class relationships across different environmental contexts.

\textbf{Decoupled Alignment (PPT-DA)} shows competitive performance with the advantage of maintaining separate optimization pathways for each platform. On MuCAR-3, it achieves the best results, suggesting that explicit decoupling may be beneficial when platform characteristics are sufficiently distinct.

\section{CONCLUSIONS}

The effectiveness of a unified framework combining PTv3 with PPT for multi-platform 3D semantic segmentation in unstructured outdoor environments is verified. This approach successfully addresses the challenges of heterogeneous LiDAR data processing through platform-specific conditioning and innovative alignment strategies. The experimental results demonstrate substantial performance improvements across diverse robotic platforms, with mIoU increases of up to 22.59\% compared to baseline approach. The framework's ability to handle domain shift and negative transfer effects while maintaining platform-specific adaptation capabilities makes it particularly suitable for field robotics applications where diverse sensor configurations and environmental contexts are common.

\bibliographystyle{IEEEtran}
\bibliography{IEEEabrv, bib}

\begin{thebibliography}{10}
\providecommand{\url}[1]{#1}
\csname url@samestyle\endcsname
\providecommand{\newblock}{\relax}
\providecommand{\bibinfo}[2]{#2}
\providecommand{\BIBentrySTDinterwordspacing}{\spaceskip=0pt\relax}
\providecommand{\BIBentryALTinterwordstretchfactor}{4}
\providecommand{\BIBentryALTinterwordspacing}{\spaceskip=\fontdimen2\font plus
\BIBentryALTinterwordstretchfactor\fontdimen3\font minus \fontdimen4\font\relax}
\providecommand{\BIBforeignlanguage}[2]{{%
\expandafter\ifx\csname l@#1\endcsname\relax
\typeout{** WARNING: IEEEtran.bst: No hyphenation pattern has been}%
\typeout{** loaded for the language `#1'. Using the pattern for}%
\typeout{** the default language instead.}%
\else
\language=\csname l@#1\endcsname
\fi
#2}}
\providecommand{\BIBdecl}{\relax}
\BIBdecl

\bibitem{goose}
\BIBentryALTinterwordspacing
P.~Mortimer, R.~Hagmanns, M.~Granero, T.~Luettel, J.~Petereit, and H.-J. Wuensche, ``{The GOOSE Dataset for Perception in Unstructured Environments},'' 2023. [Online]. Available: \url{https://arxiv.org/abs/2310.16788}
\BIBentrySTDinterwordspacing

\bibitem{gooseex}
\BIBentryALTinterwordspacing
R.~Hagmanns, P.~Mortimer, M.~Granero, T.~Luettel, and J.~Petereit, ``Excavating in the wild: The goose-ex dataset for semantic segmentation,'' 2024. [Online]. Available: \url{https://arxiv.org/abs/2409.18788}
\BIBentrySTDinterwordspacing

\bibitem{ptv3}
\BIBentryALTinterwordspacing
X.~Wu, L.~Jiang, P.-S. Wang, Z.~Liu, X.~Liu, Y.~Qiao, W.~Ouyang, T.~He, and H.~Zhao, ``Point transformer v3: Simpler, faster, stronger,'' 2024. [Online]. Available: \url{https://arxiv.org/abs/2312.10035}
\BIBentrySTDinterwordspacing

\bibitem{ppt}
\BIBentryALTinterwordspacing
X.~Wu, Z.~Tian, X.~Wen, B.~Peng, X.~Liu, K.~Yu, and H.~Zhao, ``Towards large-scale 3d representation learning with multi-dataset point prompt training,'' 2024. [Online]. Available: \url{https://arxiv.org/abs/2308.09718}
\BIBentrySTDinterwordspacing

\bibitem{minkunet}
\BIBentryALTinterwordspacing
D.~Jia and B.~Leibe, ``Person-minkunet: 3d person detection with lidar point cloud,'' 2021. [Online]. Available: \url{https://arxiv.org/abs/2107.06780}
\BIBentrySTDinterwordspacing

\bibitem{pt}
\BIBentryALTinterwordspacing
H.~Zhao, L.~Jiang, J.~Jia, P.~Torr, and V.~Koltun, ``Point transformer,'' 2021. [Online]. Available: \url{https://arxiv.org/abs/2012.09164}
\BIBentrySTDinterwordspacing

\bibitem{ptv2}
\BIBentryALTinterwordspacing
X.~Wu, Y.~Lao, L.~Jiang, X.~Liu, and H.~Zhao, ``Point transformer v2: Grouped vector attention and partition-based pooling,'' 2022. [Online]. Available: \url{https://arxiv.org/abs/2210.05666}
\BIBentrySTDinterwordspacing

\bibitem{2dpass}
\BIBentryALTinterwordspacing
X.~Yan, J.~Gao, C.~Zheng, C.~Zheng, R.~Zhang, S.~Cui, and Z.~Li, ``2dpass: 2d priors assisted semantic segmentation on lidar point clouds,'' 2022. [Online]. Available: \url{https://arxiv.org/abs/2207.04397}
\BIBentrySTDinterwordspacing

\bibitem{lcps}
\BIBentryALTinterwordspacing
Z.~Zhang, Z.~Zhang, Q.~Yu, R.~Yi, Y.~Xie, and L.~Ma, ``Lidar-camera panoptic segmentation via geometry-consistent and semantic-aware alignment,'' 2023. [Online]. Available: \url{https://arxiv.org/abs/2308.01686}
\BIBentrySTDinterwordspacing

\bibitem{DITR}
\BIBentryALTinterwordspacing
K.~A. Zeid, K.~Yilmaz, D.~de~Geus, A.~Hermans, D.~Adrian, T.~Linder, and B.~Leibe, ``Dino in the room: Leveraging 2d foundation models for 3d segmentation,'' 2025. [Online]. Available: \url{https://arxiv.org/abs/2503.18944}
\BIBentrySTDinterwordspacing

\bibitem{dinov2}
\BIBentryALTinterwordspacing
M.~Oquab, T.~Darcet, T.~Moutakanni, H.~Vo, M.~Szafraniec, V.~Khalidov, P.~Fernandez, D.~Haziza, F.~Massa, A.~El-Nouby, M.~Assran, N.~Ballas, W.~Galuba, R.~Howes, P.-Y. Huang, S.-W. Li, I.~Misra, M.~Rabbat, V.~Sharma, G.~Synnaeve, H.~Xu, H.~Jegou, J.~Mairal, P.~Labatut, A.~Joulin, and P.~Bojanowski, ``Dinov2: Learning robust visual features without supervision,'' 2024. [Online]. Available: \url{https://arxiv.org/abs/2304.07193}
\BIBentrySTDinterwordspacing

\bibitem{selfsuper}
\BIBentryALTinterwordspacing
X.~Wu, D.~DeTone, D.~Frost, T.~Shen, C.~Xie, N.~Yang, J.~Engel, R.~Newcombe, H.~Zhao, and J.~Straub, ``Sonata: Self-supervised learning of reliable point representations,'' 2025. [Online]. Available: \url{https://arxiv.org/abs/2503.16429}
\BIBentrySTDinterwordspacing

\bibitem{contrast}
\BIBentryALTinterwordspacing
M.~Afham, I.~Dissanayake, D.~Dissanayake, A.~Dharmasiri, K.~Thilakarathna, and R.~Rodrigo, ``Crosspoint: Self-supervised cross-modal contrastive learning for 3d point cloud understanding,'' 2022. [Online]. Available: \url{https://arxiv.org/abs/2203.00680}
\BIBentrySTDinterwordspacing

\bibitem{kitti}
\BIBentryALTinterwordspacing
J.~Behley, M.~Garbade, A.~Milioto, J.~Quenzel, S.~Behnke, C.~Stachniss, and J.~Gall, ``Semantickitti: A dataset for semantic scene understanding of lidar sequences,'' 2019. [Online]. Available: \url{https://arxiv.org/abs/1904.01416}
\BIBentrySTDinterwordspacing

\bibitem{nusense}
\BIBentryALTinterwordspacing
H.~Caesar, V.~Bankiti, A.~H. Lang, S.~Vora, V.~E. Liong, Q.~Xu, A.~Krishnan, Y.~Pan, G.~Baldan, and O.~Beijbom, ``nuscenes: A multimodal dataset for autonomous driving,'' 2020. [Online]. Available: \url{https://arxiv.org/abs/1903.11027}
\BIBentrySTDinterwordspacing

\bibitem{clip}
\BIBentryALTinterwordspacing
A.~Radford, J.~W. Kim, C.~Hallacy, A.~Ramesh, G.~Goh, S.~Agarwal, G.~Sastry, A.~Askell, P.~Mishkin, J.~Clark, G.~Krueger, and I.~Sutskever, ``Learning transferable visual models from natural language supervision,'' 2021. [Online]. Available: \url{https://arxiv.org/abs/2103.00020}
\BIBentrySTDinterwordspacing

\bibitem{mucar3}
M.~Himmelsbach, T.~Luettel, F.~Hecker, F.~Hundelshausen, and H.-J. Wuensche, ``Autonomous off-road navigation for mucar-3,'' \emph{KI - Künstliche Intelligenz}, vol.~25, pp. 145--149, 05 2011.

\bibitem{alice}
C.~Frese, A.~Zube, P.~Woock, T.~Emter, N.~Heide, A.~Albrecht, and J.~Petereit, ``An autonomous crawler excavator for hazardous environments,'' \emph{at - Automatisierungstechnik}, vol.~70, pp. 859--876, 10 2022.

\bibitem{adamw}
\BIBentryALTinterwordspacing
I.~Loshchilov and F.~Hutter, ``Decoupled weight decay regularization,'' 2019. [Online]. Available: \url{https://arxiv.org/abs/1711.05101}
\BIBentrySTDinterwordspacing

\bibitem{lr}
\BIBentryALTinterwordspacing
L.~N. Smith and N.~Topin, ``Super-convergence: Very fast training of neural networks using large learning rates,'' 2018. [Online]. Available: \url{https://arxiv.org/abs/1708.07120}
\BIBentrySTDinterwordspacing

\bibitem{loss}
\BIBentryALTinterwordspacing
M.~Berman, A.~R. Triki, and M.~B. Blaschko, ``The lov\'asz-softmax loss: A tractable surrogate for the optimization of the intersection-over-union measure in neural networks,'' 2018. [Online]. Available: \url{https://arxiv.org/abs/1705.08790}
\BIBentrySTDinterwordspacing

\end{thebibliography}

\end{document}